\documentclass[journal]{IEEEtran}
\usepackage{xcolor,soul,framed} 
\colorlet{shadecolor}{yellow}
\usepackage{graphicx}
\graphicspath{{../pdf/}{../jpeg/}}
\usepackage[cmex10]{amsmath}
\usepackage{array}
\usepackage{mdwmath}
\usepackage{mdwtab}
\usepackage{eqparbox}
\usepackage{url}
\usepackage{bm}
\usepackage{enumerate}
\usepackage{amsfonts}
\usepackage{nomencl}
\usepackage{etoolbox}
\usepackage{amssymb}
\usepackage{mathrsfs}
\usepackage{amsmath}
\usepackage{amsfonts,amssymb}
\usepackage{float}
\graphicspath{{figures/}}
\usepackage{calc}
\usepackage{tabularx}
\usepackage{fancybox}
\usepackage{subfig}
\usepackage{xcolor}
\usepackage[ruled,linesnumbered, longend, vlined]{algorithm2e}
\usepackage{bm}
\usepackage{bbm}

\DeclareMathOperator*{\argmax}{arg\,max}

\usepackage{pdfpages}
\usepackage{svg}
\usepackage{booktabs}
\usepackage{tabu}
\usepackage{multirow}
\usepackage{makecell}
\usepackage{array}
\usepackage{hyperref}

\begin{document}

\title{Interpretable Load Forecasting via Representation Learning of Geo-distributed Meteorological Factors}

\author{Yangze Zhou, Guoxin Lin, Gonghao Zhang, Yi Wang
}

\markboth{Submitted to IEEE Transactions on Neural Networks and Learning Systems}%
{Shell \MakeLowercase{\textit{et al.}}: Bare Demo of IEEEtran.cls for IEEE Journals}

\maketitle

\begin{abstract}
Meteorological factors (MF) are crucial in day-ahead load forecasting as they significantly influence the electricity consumption behaviors of consumers. Numerous studies have incorporated MF into the load forecasting model to achieve higher accuracy. Selecting MF from one representative location or the averaged MF as the inputs of the forecasting model is a common practice. However, the difference in MF collected in various locations within a region may be significant, which poses a challenge in selecting the appropriate MF from numerous locations. A representation learning framework is proposed to extract geo-distributed MF while considering their spatial relationships. In addition, this paper employs the Shapley value in the graph-based model to reveal connections between MF collected in different locations and loads. To reduce the computational complexity of calculating the Shapley value, an acceleration method is adopted based on Monte Carlo sampling and weighted linear regression. Experiments on two real-world datasets demonstrate that the proposed method improves the day-ahead forecasting accuracy, especially in extreme scenarios such as the ``accumulation temperature effect" in summer and ``sudden temperature change" in winter. We also find a significant correlation between the importance of MF in different locations and the corresponding area’s GDP and mainstay industry.
\end{abstract}

\IEEEpeerreviewmaketitle

\section{Introduction}

Accurate day-ahead load forecasting plays an essential role in the operation of power systems \cite{quan2013short}. Improving the accuracy of day-ahead load forecasting is an efficient approach to decreasing operational costs by reducing reserve requirements and avoiding temporary generation adjustment \cite{ranaweera1997economic,o2023attention}. Meteorological factors (MF) are crucial in day-ahead load forecasting as they significantly influence the electricity consumption behaviors of consumers. For example, the increase in temperature leads to 6.14\% and 11.3\% electricity demand in summer and spring in New South Wales, respectively \cite{ahmed2012climate}. Additionally, the relationship between electricity demand and air temperature exhibits a non-linear pattern, as demonstrated by studies conducted in London and Athens \cite{psiloglou2009factors}. These findings highlight the importance of considering MF when forecasting electricity load.

There are two primary methods for incorporating MF into the load forecasting model: used for similar daily sample selection and used as input features \cite{wang2022short}. Similar daily sample selection refers to forecasting the load for a target date with the load data from historical dates deemed similar to the target date \cite{jiang2023nsdar}. An essential aspect of this approach is developing an appropriate index for selecting similar days, and MF are typically incorporated into the index design. \cite{karimi2018priority} proposed an index for selecting similar days that involves temperature similarity and date proximity. \cite{rahman1988expert} investigated the exponential relationship between temperature and load demand and utilized a weighted mean hourly load from three previous similar days for forecasting.  \cite{gao2017weather} developed an index that considered factors such as temperature, humidity, wind speed, weather conditions, and haze to reflect human comfort. The second approach involves the MF as input for the forecasting model, allowing the model to learn the relationship between these factors and load in model training \cite{cai2019day}. It has been widely adopted to different models, such as artificial neural networks \cite{khotanzad1997annstlf,yan2012toward}, tree-based models \cite{lahouar2015day}, and support vector machines \cite{chen2017short}. 
However, weather stations are commonly built at various geographical locations within a city to monitor and record MF. Due to the urban heat island effect \cite{yang2020impact} and atmospheric activity \cite{tam2015impact}, the recorded data from different locations tends to show strong spatial heterogeneity and variability \cite{cao2021within}. Specifically, air temperature in urban areas can be higher than that in their surrounding rural areas by up to 5 $^{\circ}C$ \cite{santamouris2015analyzing}. Hence, a problem arises: MF collected from which locations should be selected for day-ahead load forecasting?

Such a problem is mainly discussed in research related to Global Energy Forecasting Competition 2012 (GEFCom2012) \cite{hong2014global} and Global Energy Forecasting Competition 2014 (GEFCom2014) \cite{hong2016probabilistic}. These competitions provided data collected by various weather stations but no identification of their geographical locations. \cite{haben2016hybrid} and \cite{mangalova2016sequence} employed simple techniques to tackle this issue, directly averaging the MF of all stations. \cite{ziel2016lasso} selected stations based on the cubic regression's in-sample fits, and the best two stations were aggregated by averaging. \cite{gaillard2016additive} utilized the generalized cross-validation criterion to select stations and created a virtual station by averaging the lowest validation loss across multiple stations. 
\cite{hong2015weather} proposed a rank-based method with three main procedures, including ranking stations based on Hongtao's vanilla benchmark \cite{hong2014global},  creating virtual stations by averaging the top weather stations, and selecting the best virtual stations based on the performance of the validation dataset. 
\cite{moreno2020rethinking} presented a genetic algorithm-based selection approach based on  \cite{hong2015weather}. The optimum weather station combination was generated by crossover and mutation during the creation of the next generation. However, current studies suffer from two main drawbacks:
\begin{enumerate}
    \item These studies neglect spatial relationships between weather stations due to the lack of geographical information. It may lead to the inappropriate merging of MF selected by stations that are far apart. In addition, different load zones may consist of various customer types, including residential, commercial, and industrial customers \cite{hong2015weather}. These different types of users are influenced differently by weather fluctuation. For instance, temperature changes have a minimal impact on the industrial load, while there is a strong impact on the residential load \cite{chen2001temperature}. Incorporating geographical information can help to capture the varying impacts of MF on load at different locations.
    \item Weather station selection and forecasting model training are treated as two separate processes in these studies. However, the impact of MF of a weather station might vary with time due to factors like commuting patterns. Specifically, MF in the downtown may be more important during working hours and the MF in non-downtown may become more crucial after work hours. To address this issue, an approach that automatically learns these relationships between all stations is necessary.
\end{enumerate}

To this end, this paper proposes a framework based on graph convolutional neural networks (GCN) to learn the meteorological representation of geo-distributed MF while considering their spatial relationship. This paper makes the following contributions:

\begin{enumerate}
    \item Studying the impact of geo-distributed MF on load forecasting, which is a rarely touched topic. This work proposes a graph convolutional neural network-based load forecasting framework that extracts meteorological representation automatically while considering spatial relationships of different locations. This approach provides more accurate day-ahead load forecasts by capturing the relationships between geo-distributed MF and load.
    \item Providing an interpretability algorithm for graph-based forecasting models using Shapley value. To address the challenge of the exponential calculation complexity associated with the Shapley value, we propose an accelerated approximation algorithm that incorporates Monte Carlo sampling and selects weighted linear regression as the surrogate model.
    \item Conducting extensive experiments on two real-world datasets. The results demonstrate that the proposed framework can achieve better forecasting accuracy in cities with more significant MF variation caused by terrain differences. Furthermore, the findings highlight a significant correlation between the importance of geo-distributed MF and the corresponding area's GDP and mainstay industry.
\end{enumerate}

The remainder of this paper is organized as follows: Section \ref{ps} defines the problem to be solved; Section \ref{methodology} designs an interpretable load forecasting method utilizing geo-distributed MF; Section \ref{case} conducts case studies on two real-world datasets; Section \ref{Conclusions} draws the conclusions.

\section{Problem Statement}
\label{ps}
Assume there are $n$ MF collection locations within the studied region, denoted as $V$. For each locations $v_i \in V$, it contains $m$ MF features $X_{i}=\{X_{i,1},X_{i,2},\cdots,X_{i,m}\}$ and their set are represented as $X=\{X_{i}|v_i \in V\}$. The geometrical relationship of $V$ is represented by the adjacency matrix $A \in \mathbb{R}^{n\times n}$. The element $A_{ij}$ denotes the geometric relationship between locations $i$ and $j$. $A_{ij}=1$ means that they are adjacent; $A_{ij}=0$ means they are not adjacent. 

Traditionally, MF of location $v_i$ are selected or extracted before model training. The pre-selected or pre-extracted MF (denoted as $X'$), together with other features (such as calendar variables and history loads, denoted as $X_O$), are fed as inputs to the forecasting model $f$. Then, the model is trained as:
\begin{equation}
    \arg \min_{f} \sum_{t\in T_{\text{train}}} L(f(X'_t,X_{O,t}),y_t)
\end{equation}
where $T_{\text{train}}$ is the index set of training dataset and $X'_t$, $,X_{O,t}$, and $y_t$ are $t$-th pre-selected or pre-extracted MF, other feature, and actual load, respectively.

Traditional approaches treat feature selection and model training as two separate processes. 
Here, we propose an integrated load forecasting model to combine the representation generation model $g$ and the load forecasting model $f$. In addition, to fully consider geo-distribution, the adjacency matrix $A$ is considered by $g$ to extract the representation of geo-distributed MF. This process can be represented as:
\begin{equation}
\label{eq_ps}
    \arg \min_{g,f} \sum_{t\in T_{\text{train}}} L(f(g(X_{t},A),X_{O,t}),y_t)
\end{equation}

Based on \eqref{eq_ps}, we focus on the following two tasks:
\begin{enumerate}
    \item Developing an integrated load forecasting model to extract the representation of geo-distributed MF automatically to produce higher accurate forecasts;
    \item Interpreting the complex forecasting model to underlie the relationship between MF and loads to 
    provide practical guidance on deploying MF collection stations.
\end{enumerate}

\begin{figure*}[htbp]
\centering
\includegraphics[scale=0.46]{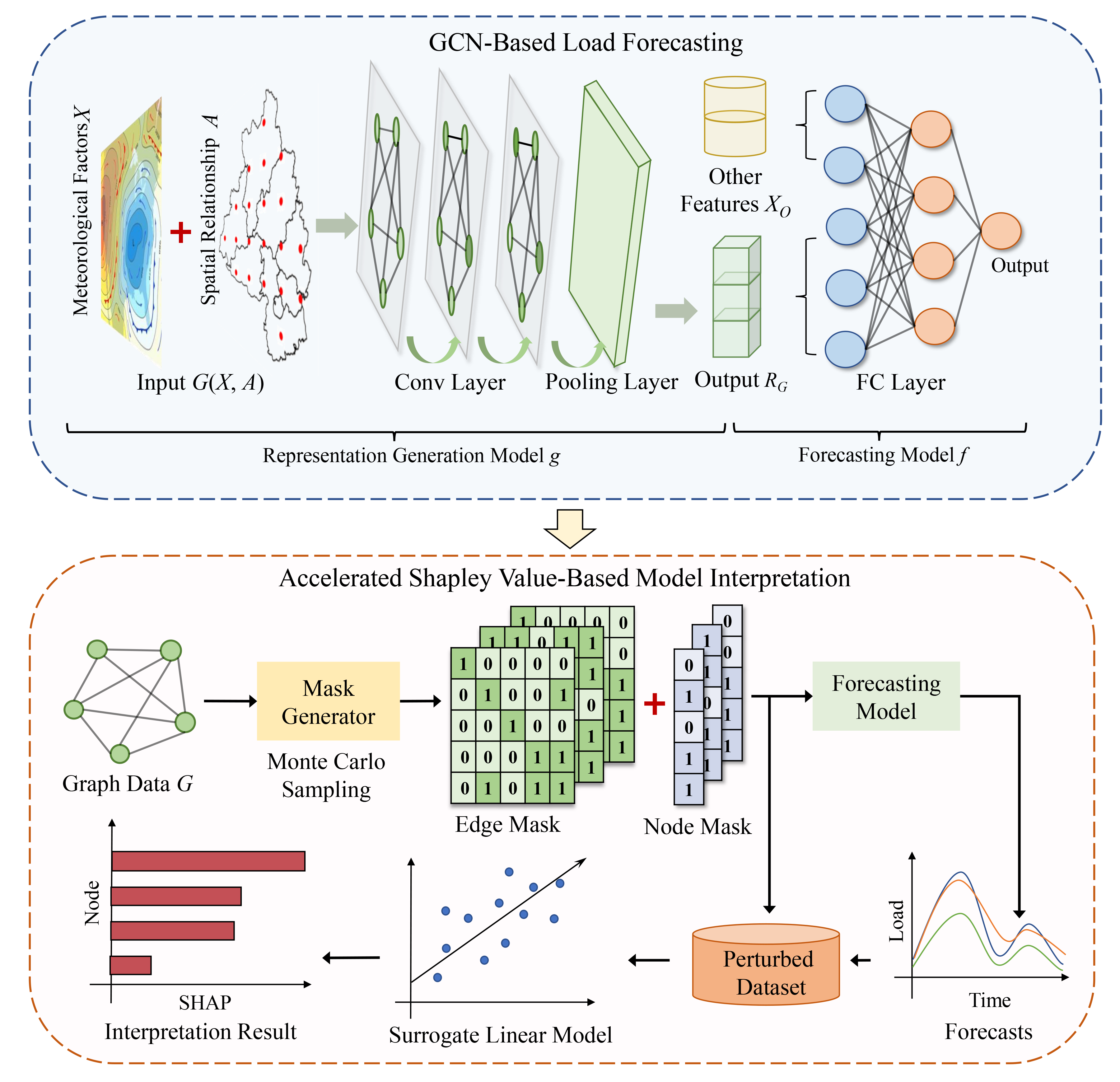}
\caption{The proposed load forecasting framework utilizing geo-distributed MF.}
\label{framework}
\end{figure*}

\section{Proposed Framework}
\label{methodology}
Fig. \ref{framework} presents the proposed load forecasting framework using geo-distributed MF. It consists of two procedures, i.e., GCN-based load forecasting and accelerated Shapley value-based model interpretation, which correspond to the two tasks defined above.

\subsection{GCN-based Load Forecasting}
GCN is one of the most popular variants of graph neural networks (GNN), which is specifically designed to operate on graph-structured data \cite{kipf2016semi}. It has been successfully applied in extracting hidden information from diverse types of graph data, including traffic networks \cite{zhao2019t} and social networks \cite{yu2020enhancing}. For applications in energy systems, GCN has demonstrated its capability in integrating network topology information and temporal dynamics into voltage disturbance detection models \cite{luo2021data}. It has been utilized to describe the relationships among different blocks within complex buildings, leading to improved energy consumption forecasting \cite{lu2022graph}. 

Here, we focus on how to extract the representation of geo-distributed MF to improve load forecasting accuracy with GCN. The proposed GCN-based load forecasting can be divided into representation generation model $g$, load forecasting model $f$, and integrated model training.

The representation generation model gg consists of an input layer, convolutional layers, pooling layers, and an output layer. The input is graph $G(X,A)$ composed of geo-distributed MF and the output is the extracted representation $R_G$. The function of the convolutional layer can be divided into three steps: propagation, aggregation, and updating. In the propagation step, the model computes messages between every node pair $(v_i,v_j)$. In the aggregation step, for each $v_i \in V $in the graph, the messages from its neighboring nodes are aggregated into $m^{(l-1)}_i$, where $(l)$ is the ll-th convolutional layer. In the updating step, the updated representation $H_i^{(l)}$ of each node $v_i$ is computed using the aggregated messages $m_i^{(l-1)}$ and the previous layer representation $H_i^{(l-1)}$. The working mechanism of convolutional layers can be expressed as \eqref{gcn} \cite{kipf2016semi}. The pooling layer is adopted in the last convolutional layer, which reduces the dimension of the graph representation by aggregating the representation of the last convolutional layer.
\begin{equation}
\centering
    \begin{split}
    H^{(l)}=\sigma(\widetilde{D}^{-\frac{1}{2}}\widetilde{A}\widetilde{D}^{-\frac{1}{2}}H^{(l-1)}\Theta_g^{(l)}) \\
    \widetilde{A}=A+I,   ~  \widetilde{D}=\sum_j \widetilde{A}_{ij}
\end{split}
\label{gcn}
\end{equation}
where $I$ is identity matrix, $D$ is a degree matrix, $\sigma(\cdot)$ is activation function, $\Theta_g^{(l)}$ is the trainable weights $l$-th layer of $g$, $H^{(l)}$ is the representation of $l$-th layer.

The output of the representation generation model $R_G$, together with other features $X_O$, are fed as inputs to the forecasting model $f$. Here, $g$ and $f$ are trained simultaneously using back-propagation algorithms and the chain principle. For the $q$-th training epoch. the gradient of the loss function over $R_G$ can be back-propagated to update $g$:
\begin{equation}
    \begin{split}
        \Theta_f(q)&=\Theta_f(q-1)-\lambda \frac{\partial L}{\partial \Theta_f}\\
    \Theta_g(q)&=\Theta_g(q-1)-\lambda \frac{\partial L}{\partial R_G} \frac{\partial R_G}{\partial \Theta_g}
\end{split}\label{training}
\end{equation}
where $\lambda$ is learning rate, $L$ is the forecasting error, and $\Theta_g$, and $\Theta_f$ are trainable parameters of $g$ and $f$. 

To prevent overfitting, an early-stopping strategy is employed, i.e., the training process will be stopped if the loss of $f$ on the validation dataset does not decrease for a certain number of training epochs.

\subsection{Accelerated Shapley Value-Based Model Interpretation}
\begin{figure*}[htbp]
\centering
\includegraphics[scale=0.50]{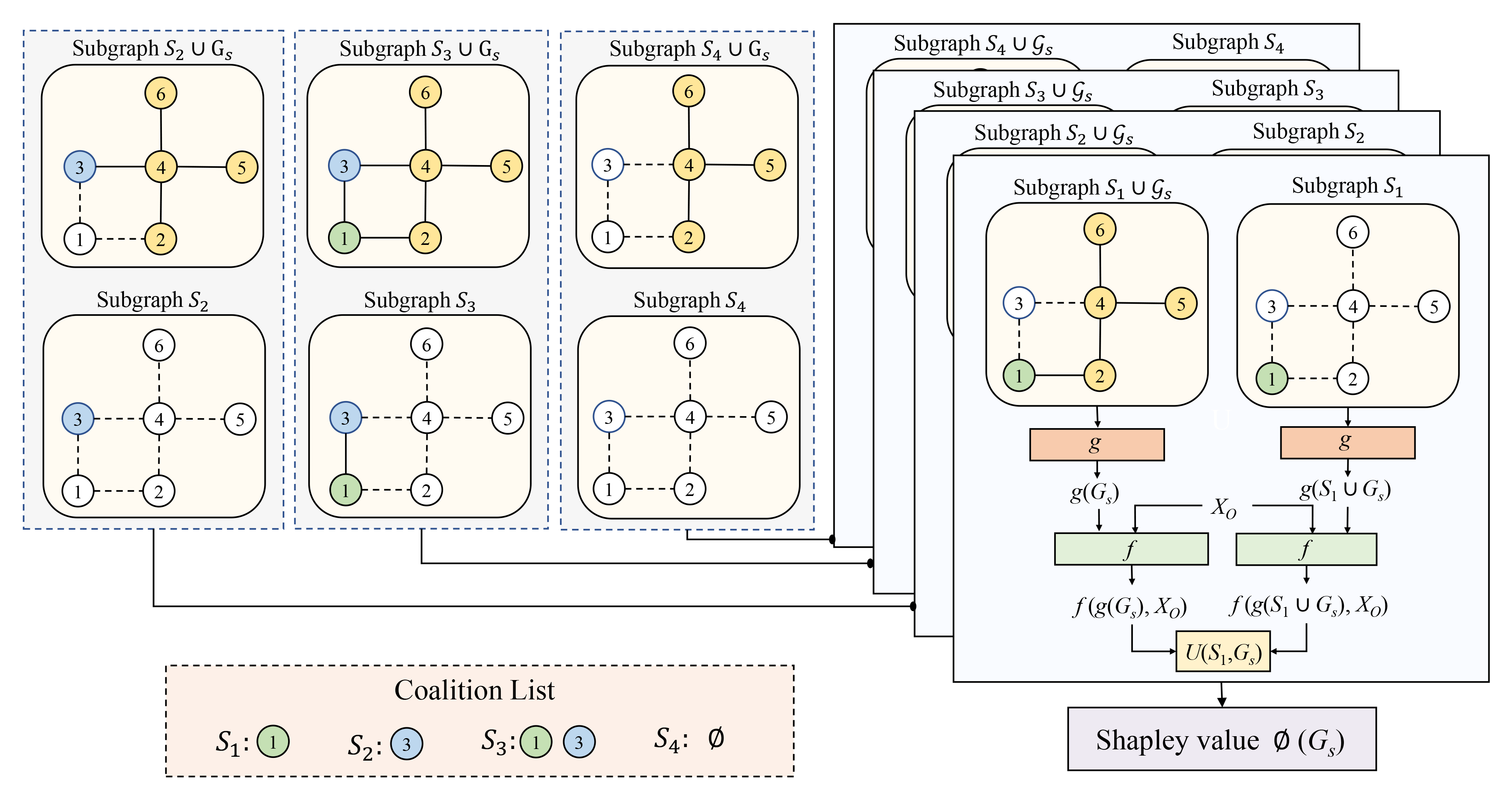}
\caption{An illustration of Shapley value for interpreting GNN}
\label{shap illustration}
\end{figure*}
Given the presence of numerous MF collection locations within a region, it is necessary to interpret the forecasting model to determine MF from which locations are more important. It can underly the connections between geo-distributed MF and loads, guiding practical weather station deployment.

The goal of model interpretation is to find the most important subgraph $G_s(X_s, A_s)$ for the forecasts $y$, which can be formulated as \eqref{intepretation}.
\begin{equation}
    G_s=\argmax_{|G_s|\leq N_{\max}} \text{Score}(f \odot g,G(X,A),G_s(X_s,A_s))
    \label{intepretation}
\end{equation}
where $\text{Score}()$ is a scoring function to evaluate the importance of a subgraph, $f \odot g$ is the integrated model of $f$ and $g$, $G_s(X_s, A_s)$ is a subgraph of $G(X, A)$, $N_{max}$ is the upper bound of the size of the subgraph, which is defined as the number of nodes in this work.

\subsubsection{Shapley Value}
The Shapley value is a concept in cooperative game theory that describes how to fairly distribute the total gains of groups to participants based on their contribution \cite{duval2021graphsvx}. It has been widely used for model interpretation due to its game-theoretical properties \cite{chau2022rkhs}. It has been successfully applied to various model structures, including tree-based models \cite{moon2022toward} and neural networks \cite{zhou2022elastic}. There is also some research exploring how the Shapley value can be used to interpret graph neural networks. For instance, \cite{duval2021graphsvx} proposed a unified framework called GNN Explainer, which constructed a surrogate model on a perturbed dataset, achieving the employment of the Shapley value to GNN. However, existing works mainly focus on clustering tasks or node feature forecasting tasks. 
This work adopts the Shapley value as the score function in \eqref{intepretation} to interpret the graph-based load forecasting model where GCN serves as the representation generation model.

Given $G$ and trained integrated model $f\odot g$, we study the Shapley value of a target subgraph $G_s$, denoted the nodes in $G_s$ as $V_s$ while other nodes as $V \backslash V_s$. Shapley value considers the marginal contribution of each participant in different cooperative coalitions as defined in \eqref{shap_eq}. 
\begin{equation}\label{shap_eq}
\begin{split}
\phi(G_s)&=\sum_{S \subseteq V \backslash V_s} \frac{|S|!(n-|S|-1)!}{n!} U(S,G_s)\\
U(S,G_s)&=f(g(S\cup G_s),X_O)-f(g(G_s),X_O)
\end{split}
\end{equation}
where $\phi(G_s)$ is the Shapley value of the $G_s$, coalition size $|S|$ is the number of nodes in $S$, $ g(G_s)$ and $ g(G_s \cup S)$ means the output of $g$ when given $G_s$ and $G_s \cup S$, $U(S,G_s)$ is marginal contribution of $G_s$ given $S$.

Fig. \ref{shap illustration} illustrates how to score the importance of the subgraph $G_s$, which consists of nodes $\{2,4,5,6\}$. There are four coalitions of the nodes in $V\backslash V_s=\{1,3\}$, which are $\{1\}$,$\{3\}$,$\{1,3\}$,$\{\varnothing\}$. The difference between the output of the $f\odot g$ with the input of $S\cup G_s$ and $G_s$ for each coalition $S$ will be calculated as the marginal contribution. The Shapley value of $G_s$ is computed with the marginal contribution of $U(S, G_s)$ when given different coalitions $S$.

\subsubsection{Acceleration Strategy}
Although Shapley value has been proven to possess game-theoretical properties,  it has a drawback of exponential computational complexity. When the number of nodes in a graph is large, the computation time becomes unacceptable. There are various research to accelerate the calculation of Shapley value and the thought of Kernel Shapley value \cite{lundberg2017unified} is introduced in this work. The algorithm of the proposed accelerated graph-based model interpretation method 
is shown in the Algorithm. \ref{alg1}.

\textbf{Mask generation:}
Mask generator samples the subgraph $G_s$ of the original graph $G$ by Monte Carlo sampling. 
Specifically, for a given graph $G$, we randomly create a binary vector $M_N \in \mathbb{R}^{n}$ as the node mask. Element in $M_N$ represents whether a node is selected in the subgraph. $M_{N,i}=1$ means node $v_i \in V_s$ and $M_{N,i}=0$ means node $v_i \in V\backslash V_s$.
Then, the edge mask $M_E \in \mathbb{R}^{n\times n}$ is generated based on $M_N$. If $v_i \in V\backslash V_s$, it is not adjacent to any other nodes, implying that $M_{E,ij}=0$ for $j\neq i$. Repeat the above steps to generate $P$ pairs of $M_N$ and $M_E$, and their set are denoted as $\mathcal{M}_N=\{M_{N,p}|p\in [1,P]\}$ and $\mathcal{M}_E=\{M_{E,p}|p\in [1,P]\}$.

\textbf{Perturbed dataset generation:} 
The perturbed dataset is constructed based on the mask sets $\mathcal{M}_N$ and $\mathcal{M}_E$. These masks are applied to create a subgraph set $\mathcal{G}_s$, which is then sent to the proposed model $f\odot g$ to generate forecasts. To form the adjacent matrix $A_{s,p}$ of a subgraph $G_{s,p}$, element-wise multiplication is performed between the original adjacent matrix $A$ and $M_{E,p}$. The node features $X_{s,p}$ of $G_{s,p}$ are obtained by setting the features of nodes in $V \backslash V_{s,p}$ to zero. These generated subgraphs $\mathcal{G}_s=\{G_{s,p}|p\in [1,P]\}$ are then used as input to $g$ to extract geo-distributed MF representations, which are concatenated with $X_O$ and provided as input to the model $f$ to generate forecasts $\mathcal{Y}_s=\{\hat{y}_{s,p}| \hat{y}_{s,p}=f(g(G_{s,p}),X_O), p\in [1,P]\}$. $(\mathcal{M}_N,\mathcal{Y}_s)$ forms the perturbed dataset for the explanation generation.

\textbf{ Explanation generation:}
Kernel Shapley value adopts a linear model $h$ as the surrogate model, where the sum of the Shapley values gives the model prediction. For given subgraph $G_s$, the output of $h$ is defined:
\begin{equation}
    h(G_s)=\phi_0+\sum_{i=1}^n \phi_i M_{N,i}
\end{equation}
where $\phi$ is the trainable parameter of $h$, $M_{N}$ is the corresponding node mask of $G_s$, $\phi_0$ is the forecasts of $f \odot g$ when $V_s=\varnothing$.

The surrogate model $h$ is constructed in the approach of weighted least squares. 
The weights $\pi_s$ of the sample $(M_N,y_{s})$ in the perturbed dataset can be calculated as \eqref{weights}. These sample weights depend on the coalition size and attach more importance to samples with small or large coalition sizes to reflect the individual effect \cite{akkas2024gnnshap}.
\begin{equation}
    \pi_s=\frac{n-1}{\left(\begin{array}{c}
n \\
n_s
\end{array}\right)
    n_s(n-n_s)}
\label{weights}
\end{equation}
where $\pi_s$ is the weights of sample $(M_N,y_{s})$, $n_s=\sum_{i=1}^n M_{N,i}$ is the size of the coalition.

The training process of $h$ can be formulated as the optimization problem in \eqref{surrogate model}. 
\begin{equation}
\min \sum_{p=1}^{P} [\hat{y}_{s,p}-h(G_{s,p})]^2\pi_s
\label{surrogate model}
\end{equation}
The solution of this optimization problem is:
\begin{equation}
    \Phi = (\mathcal{M}_N^T \Pi \mathcal{M}_N)^{-1}\mathcal{M}_N \Pi \mathcal{Y}_s
\end{equation}
where $\Pi$ is the weights of all samples in $(\mathcal{M}_N,\mathcal{Y}_s)$, $\Phi$ is the trainable parameters of $h$.

\SetAlgoVlined
\begin{algorithm}[t]
\caption{Accelerated Graph-based Model Interpretation Algorithm}\label{alg1}
\SetKwInOut{KIN}{Input}
\SetKwInOut{KOUT}{Output}
\KIN{Graph $G=(X,A)$, representation generator $g$, forecasting model $f$,number of samples $P$}
\KOUT{Shapley values for all nodes in $V$}
\tcp{Mask Generation}
Randomly generates $\mathcal{M}_N$ and $\mathcal{M}_E$\;
\tcp{Perturbed Dataset Generation}
\For{$M_{N,p}$ and $M_{E,p}$ in $\mathcal{M}_N$ and $\mathcal{M}_E$}{
Create $G_{s,p}=(X_{s,p},A_{s,p})$\;
$\hat{y}_{s,p}=f(g(G_{s,p}),X_O)$\;
Create sample weight $\pi_{s,p}$\;
}
Generate perturbed dataset $(\mathcal{M}_N,\mathcal{Y}_s)$ and $\Pi$

\tcp{Explanation Generation}
$ \Phi = (\mathcal{M}_N^T \Pi \mathcal{M}_N)^{-1}\mathcal{M}_N \Pi \mathcal{Y}_s$\;

\textbf{Return $\Phi$} 
\end{algorithm}

\section{Case Studies}
\label{case}
\subsection{Experimental Setups}
\subsubsection{Data source}
This study evaluates the proposed framework using data collected from two cities in China (City A and City B). The hourly electricity load data from City A is collected from 2018 to 2023, with the data from 2018 to 2022 selected as the training dataset and the data collected in 2023 used as the testing dataset. The hourly electricity load from City B is collected from 2014 to 2018, with the data from 2014 to 2017 selected as the training dataset and the data collected in 2018 used as the testing dataset.

The MF utilized in this work are downloaded from the fifth-generation European Centre for Medium-Range Weather Forecasts (ERA5) for the global climate and weather. ERA5 provides global climate and weather data, which are gridded to a regular latitude-longitude grid of 0.25 degrees. This dataset is obtained through data assimilation, which combines model data with observations from around the world to create a global dataset using the laws of physics. There are 18 grid points in City A and 20 grid points in City B. The MF considered in this study includes the 2-meter (m) temperature and relative humidity. Furthermore, historical load features (last seven days) and calendar features (month, weekday, hour) are also chosen as load forecasting inputs as well.

\subsubsection{Metric}
Mean absolute error (MAE) and  Mean absolute percentage error (MAPE) are chosen to evaluate forecast accuracy in this work. They are defined by \eqref{mae} and \eqref{mape}.
\begin{subequations}
\begin{equation}\label{mae}
   \text{MAE}=\frac{1}{|T|}\sum_{t\in T}|y_t-\hat{y}_t|
\end{equation}
\begin{equation}\label{mape}
   \text{MAPE}=\frac{1}{|T|}\sum_{t\in T}|\frac{y_t-\hat{y}_t}{y_i}|
\end{equation}
\end{subequations}
where $y_t$ and $\hat{y}_t$ is the $t$-th actual load and predicted load, respectively; $T$ is the dataset index set, $|T|$ is the size of $T$.

In the real world, system operators tend to focus more on load forecasting accuracy at 11:00 AM and 8:00 PM. This is because the load at these times is generally higher or the forecasting errors are often larger than other periods. Hence, system operators consider these time points to be more critical for day-ahead scheduling. Therefore, the forecasting error (MAE and MAPE) at 11:00 AM and 8:00 PM will be calculated, which are denoted as $\text{MAE}_{\text{Noon}}$, $\text{MAE}_{\text{Night}}$, $\text{MAPE}_{\text{Noon}}$ and $\text{MAPE}_{\text{Night}}$. Additionally, this work proposes comprehensive metrics $\text{MAPE}_{\text{com}}$ and $\text{MAE}_{\text{com}}$ to combine the forecasting accuracy at different periods. This approach aims to provide a more comprehensive evaluation of the model's performance.
\begin{subequations}
\begin{equation}
    \text{MAE}_{\text{com}}=0.6 \text{MAE}+0.2 \text{MAE}_{\text{Noon}}+0.2 \text{MAE}_{\text{Night}}
\end{equation}
\begin{equation}
    \text{MAPE}_{\text{com}}=0.6 \text{MAPE}+0.2 \text{MAPE}_{\text{Noon}}+0.2 \text{MAPE}_{\text{Night}}
\end{equation}
\end{subequations}

\subsubsection{Benchmark}
There are several kinds of benchmarks and all forecasting models of these benchmarks are artificial neural networks. 
The first kind of benchmark is selecting MF from one location as the model input. It is denoted as $\text{L}x$, where $x$ is the index of location. As for other benchmarks,
``None MF" means the forecasting model does not accept the MF as input features.
``All locations" means that the MF from all locations is fed directly into the forecasting model.
``Average" means the geo-distributed is aggregated by taking the average and then provided to the forecasting model. ``HT" means the Hongtao weather station selection mentioned in \cite{hong2015weather}.


\subsection{Forecasting Performance}

\subsubsection{City A}
\begin{figure}[t]
\centering
\includegraphics[scale=0.6]{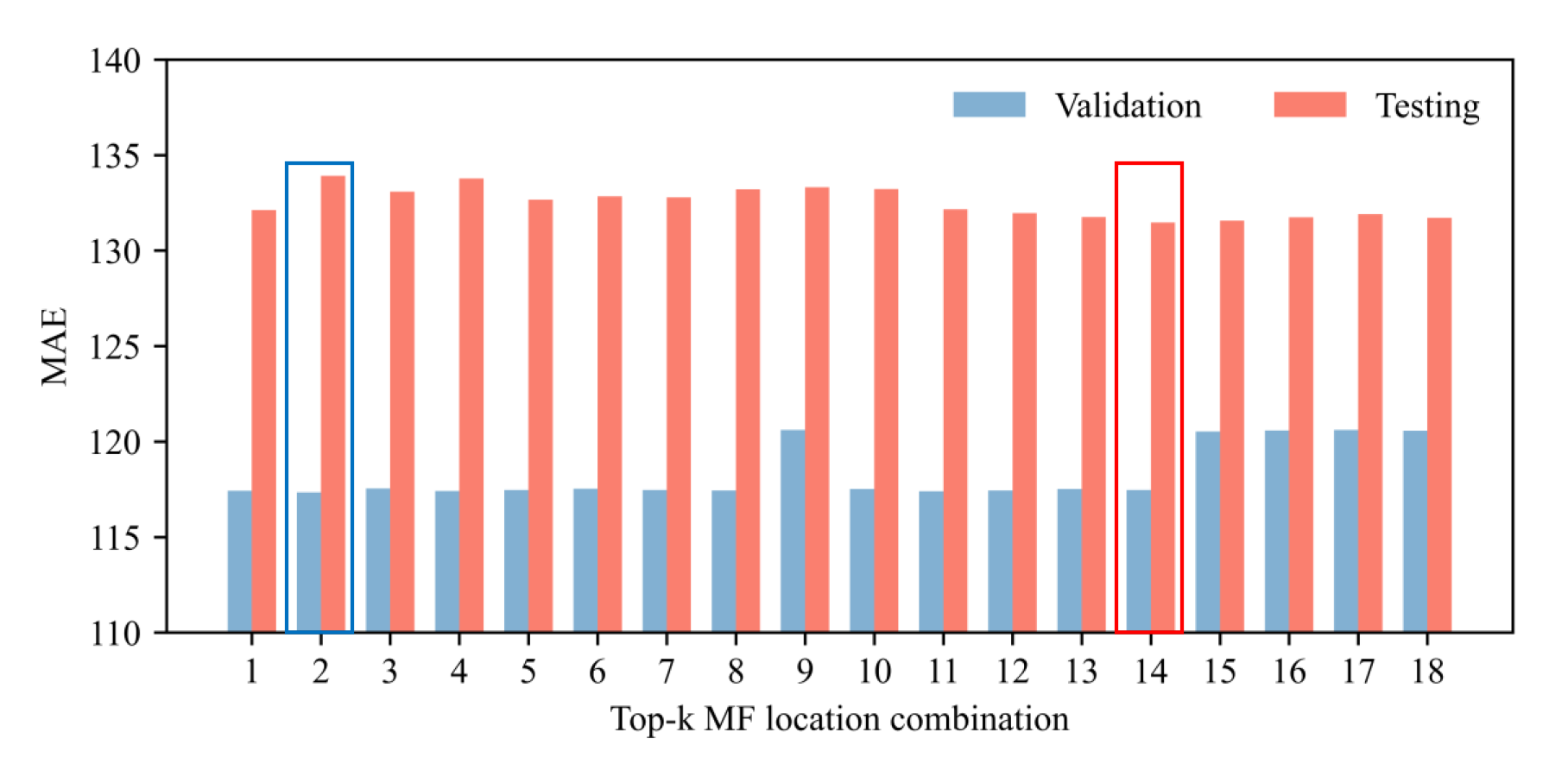}
\caption{The MAE performance of Hongtao method in City A}
\label{HT}
\end{figure}

The performance ranking of the forecasting model in the training dataset is as follows: L6, L1, L9, L2, L11, L5, L8, L3, L4, L12, L13, L10, L14, L15, L16, L7, L0, L17. 
Figure \ref{HT} displays the MAE of the validation and testing datasets when the model is provided with the average of the MF from the top-$k$ locations (where $k$ is the number of selected locations).
The blue and red boxes in the figure indicate the location combinations that result in the lowest MAE in the validation and testing datasets, respectively. It can be observed that aggregating the top 2 locations (L6 and L1) yields the best performance in the validation dataset and it is regarded as the best MF combination according to \cite{hong2015weather}. However, the best performance in the testing dataset is achieved by aggregating the top 14 locations (L6 to L15), which differs from the Hongtao approach's result.

The accuracy of different models is shown in Table. \ref{table1}. The proposed method achieves a performance enhancement of 1.722\% and 1.894\% on MAPE and MAE respectively compared to the best performance of 22 benchmarks. Specifically, the accuracy of the proposed method improves by 2.150\% and 2.028\% on $\text{MAPE}_\text{Noon}$ and $\text{MAE}_\text{Noon}$. At night, the forecasting error of the proposed method decreases significantly, with a reduction of 4.144\% on $\text{MAPE}_\text{Night}$ and 11.006\% on $\text{MAE}_\text{Night}$. As for comprehensive performance, the proposed method demonstrates a 3.624\% and 2.431\% improvement in accuracy on MAPE and MAE respectively.

\begin{figure*}[t]
\centering
\includegraphics[scale=0.7]{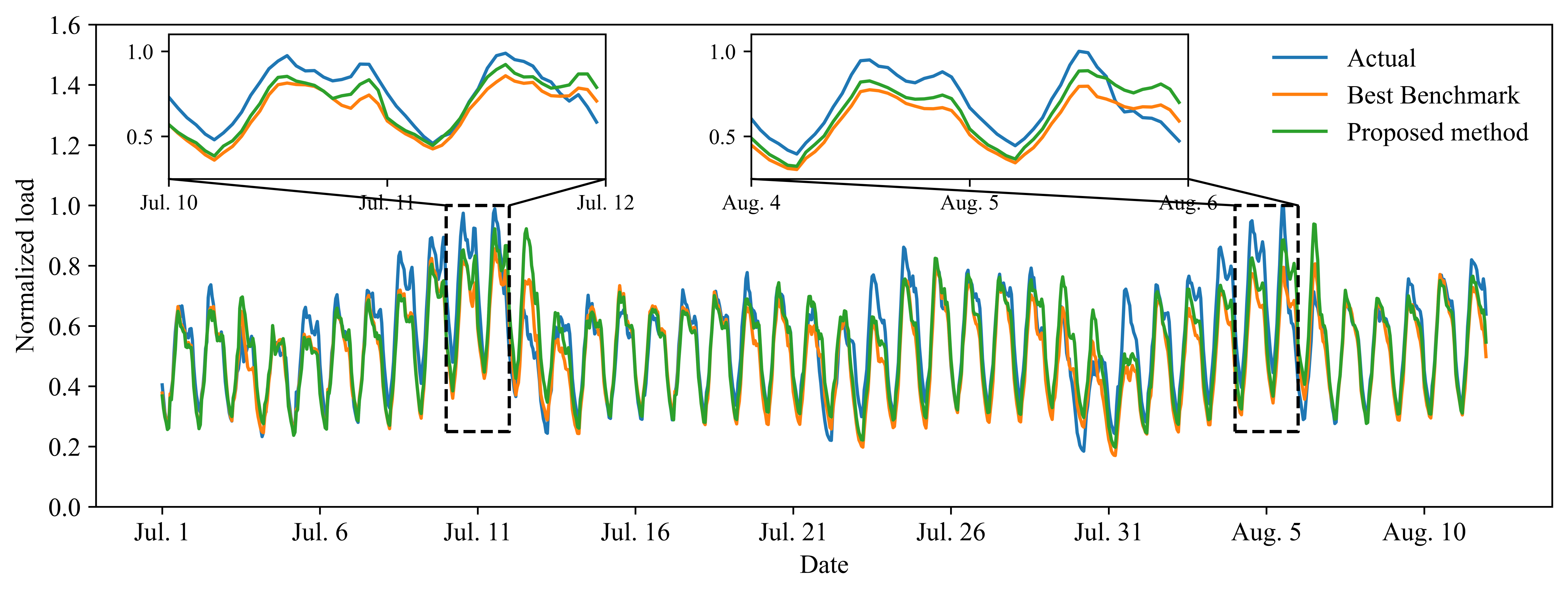}
\caption{Normalized forecasts of City A from Jul. 1 to Aug. 10}
\label{curveA}
\end{figure*}

\begin{figure}[t]
\centering
\includegraphics[scale=0.45]{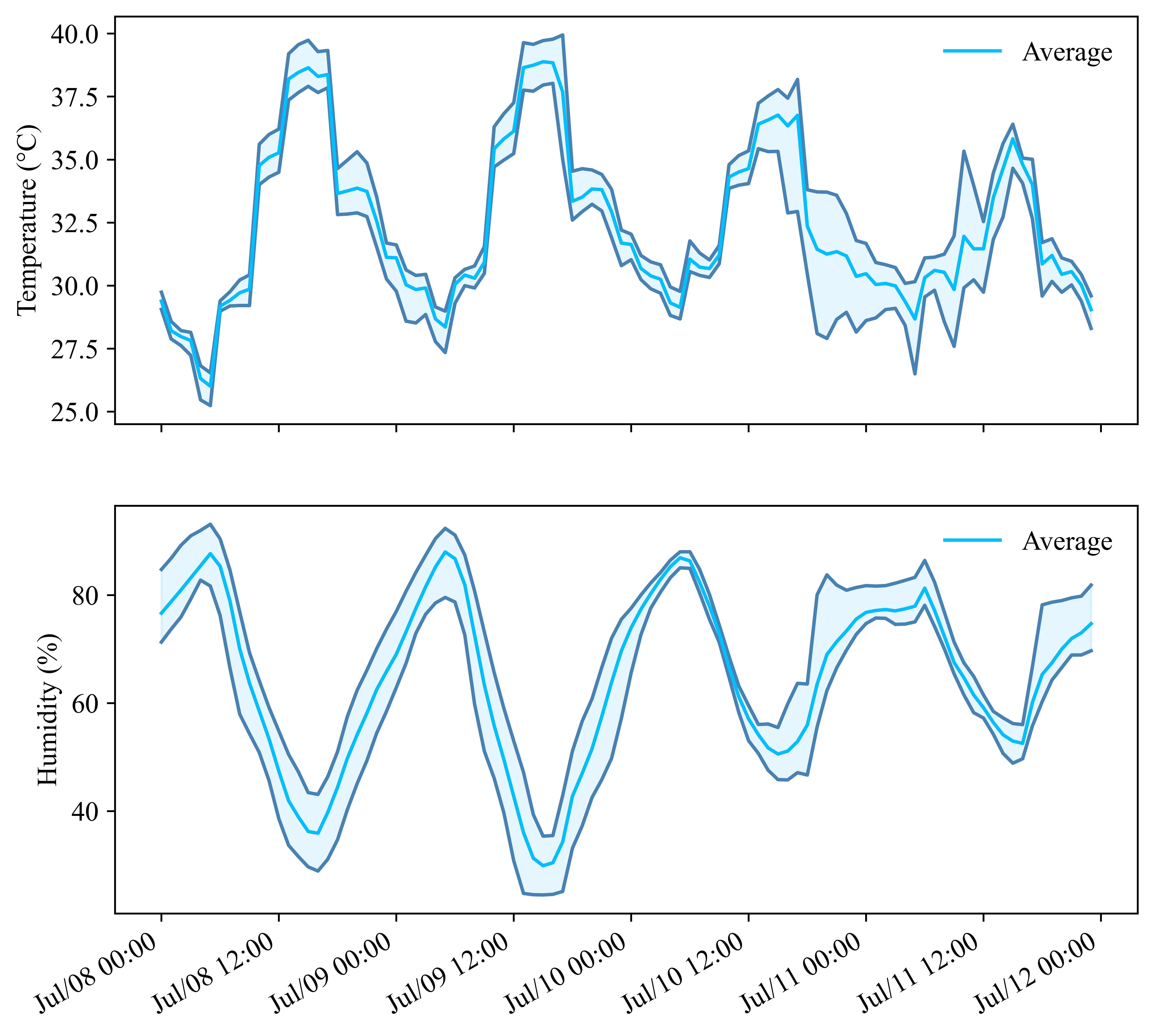}
\caption{The range of temperature and humidity of 18 locations in City A from Jul. 8 to Jul. 12}
\label{temperature humidity a}
\end{figure}

The normalized forecasts of the best benchmarks and the proposed method of Jul. 1 to Aug. 10 2023 are shown in Fig \ref{curveA}. 
Load increases significantly from Jul. 8 to Dec. 12 and Aug. 2 to Aug. 6, which is called the ``accumulation temperature effect". This effect refers to the phenomenon of higher peak electricity loads caused by prolonged high temperatures \cite{li2015modification}. In these extreme scenarios of prolonged high temperatures, our method effectively addresses the problem of under-forecasting, reducing considerable operation costs. For details, the range of temperature and humidity of 18 locations are shown in Fig. \ref{temperature humidity a}. Although the highest temperatures from Jul. 10 to Jul. 11 are lower than those from Jul. 8 to Jul. 9, the peak loads on these two days are much greater than those on the previous two days. Traditional approaches that merely consider MF from one location or some representative locations struggle to forecast the increasing load. Additionally, the differences in temperature and humidity across various locations are larger than on other days, particularly from the afternoon of Jul. 10 to the noon of Jul. 11. Correspondingly, our proposed model performed significantly better than the benchmark within this interval. 
This suggests that our method's superiority is more evident when MF from different locations shows stronger spatial heterogeneity.

\subsubsection{City B}
\begin{figure}[t]
\includegraphics[scale=0.6]{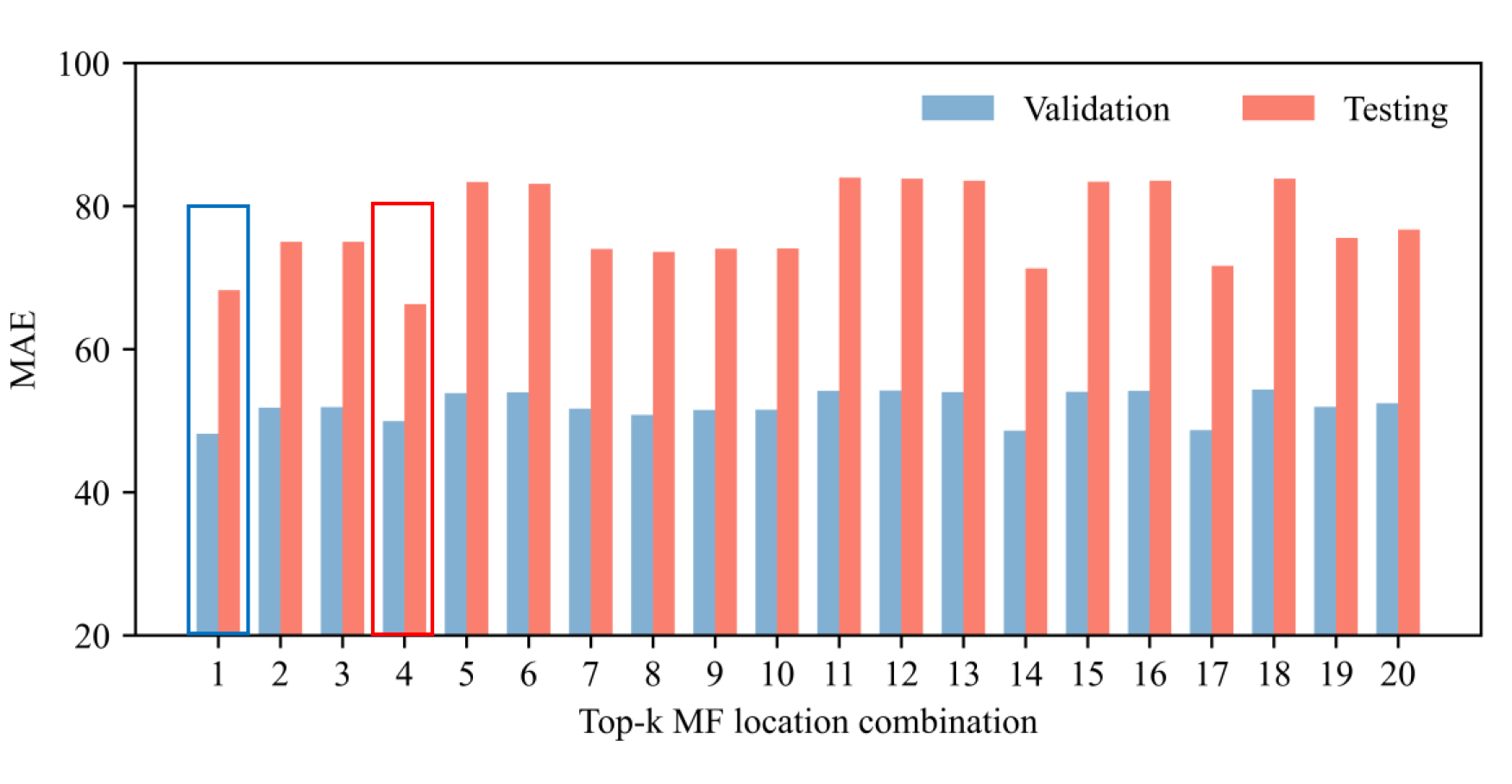}
\caption{The MAE performance of Hongtao method in City B}
\label{HT b}
\end{figure}

Table \ref{table2} shows the accuracy of different models for City B.
The performance ranking of the forecasting model in the training dataset is as follows: L15, L16, L17, L2, L11, L10, L5, L0, L8, L18, L14, L19, L4, L6, L3, L1, L9, L7, L12, L13. The performance in the validation and testing dataset of different top-$k$ location combinations is shown in Fig. \ref{HT b}. It shows that the top 1 is the best combination, which means the result of the Hongtao method is the same as L15. However, the best testing performance is obtained by the top 4 (L15 to L2) location combinations.

Compared to the best performance of 25 benchmarks, our method reduces 6.085\% and 7.873\% on MAPE and MAE, respectively. Our method reduces MAPE and MAE by 7.731\% and 8.199\% at 11:00 AM and reduces forecasting error by 6.377\% and 7.238\% at 8 PM. Overall, the proposed method demonstrates a comprehensive improvement in accuracy, with a 6.479\% and 8.165\% enhancement on MAE and MAPE. Different from case A, the accuracy improvements at these timestamps are similar to other timestamps and this phenomenon will be discussed in the analysis section. 

\begin{figure*}[t]
\centering
\includegraphics[scale=0.7]{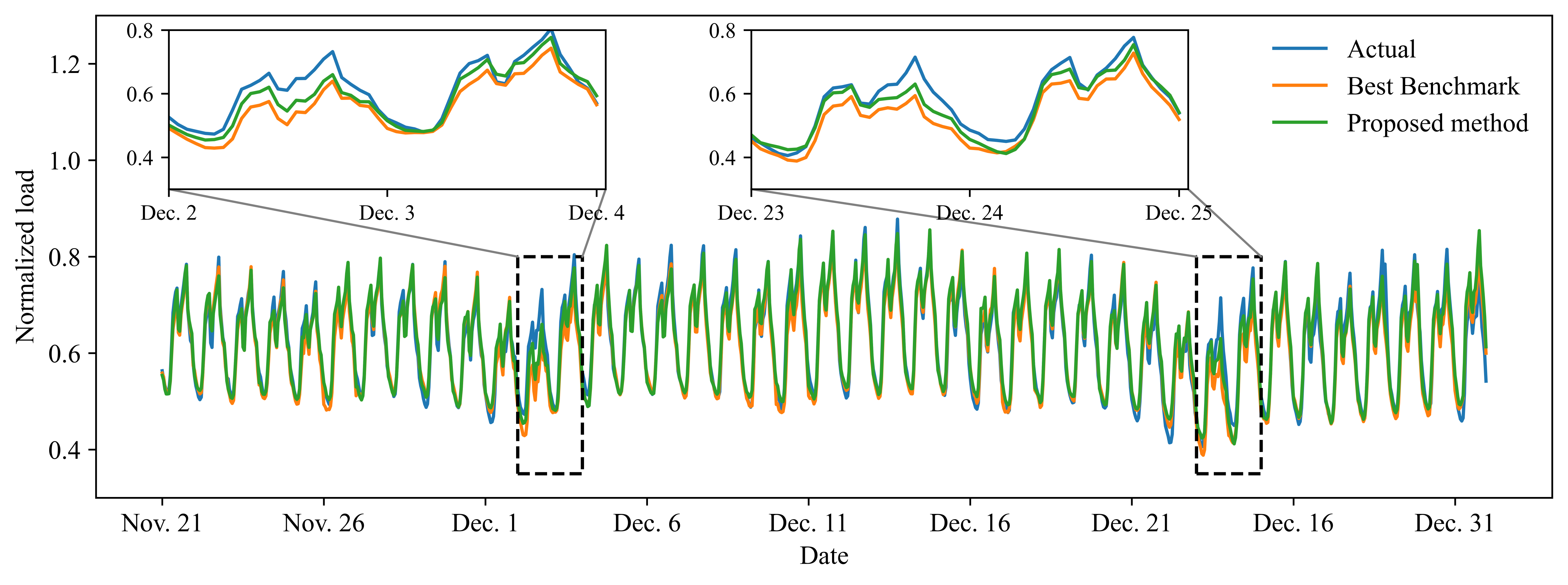}
\caption{Normalized forecasts of City B from Nov. 21 to Dec. 31 2018}
\label{curveB}
\end{figure*}

\begin{figure}[t]
\centering
\includegraphics[scale=0.45]{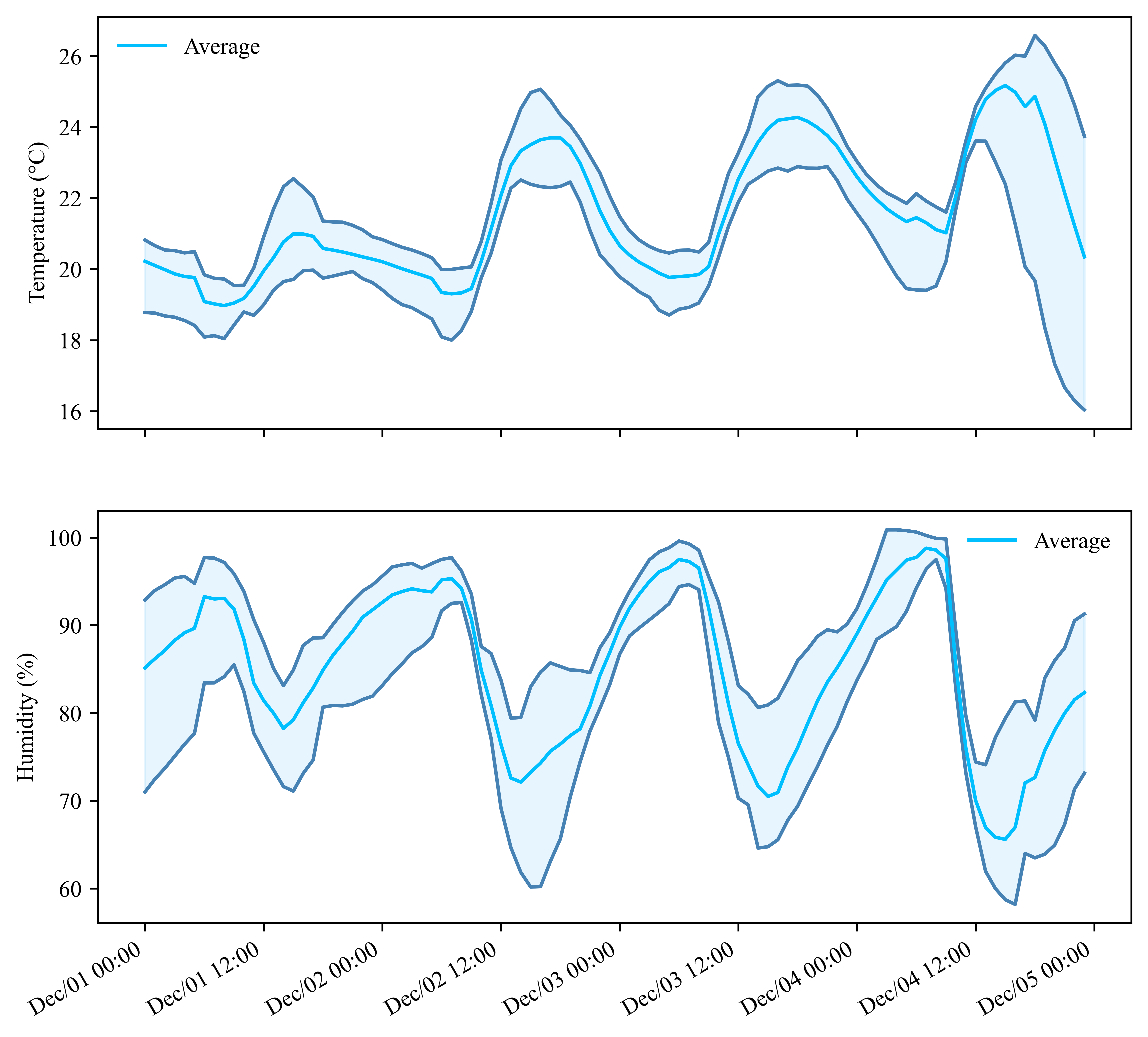}
\caption{The range of temperature and humidity of 20 locations in City B from Dec. 1 to Dec. 5, 2018}
\label{temperature humidity b}
\end{figure}

The normalized forecasts of the best benchmarks and the proposed method from Nov. 21 to Dec. 31 are shown in Fig. \ref{curveB}. 
It can be observed that there were significant load fluctuations caused by ``sudden temperature change" from Dec. 2 to Dec. 4 and Dec. 23 to Dec. 25, during which the performance of the best benchmarks decreased seriously. The range of temperature and humidity of 20 locations are illustrated in Fig. \ref{temperature humidity b}. The temperature has suddenly increased from Dec. 2, causing the concept shift of loads. 
However, our proposed method effectively mitigates the performance degradation by extracting the representation of geo-distributed MF.

\begin{table*}[t]
\centering
\caption{Accuracy of different methods in City A}
\label{table1}
\begin{tabular}{lcccccccc}
\hline
     & \multicolumn{1}{l}{MAPE} & \multicolumn{1}{l}{$\text{MAPE}_{\text{Noon}}$} & \multicolumn{1}{l}{$\text{MAPE}_{\text{Night}}$} & \multicolumn{1}{l}{$\text{MAPE}_{\text{Com}}$} & \multicolumn{1}{l}{MAE}       & \multicolumn{1}{l}{$\text{MAE}_{\text{Noon}}$}   & \multicolumn{1}{l}{$\text{MAE}_{\text{Night}}$}    & \multicolumn{1}{l}{$\text{MAE}_{\text{Com}}$}    \\ \hline
L0   & 3.631\%                  & 3.691\%                     & 3.705\%                     & 3.658\%                     &  134.94 & 155.57 & 147.71 & 141.62\\
L1   & 3.604\%                  & 3.822\%                     & 3.608\%                     & 3.648\%                     &  135.82 & 162.30 & 145.93 & 143.14\\
L2   & 3.619\%                  & 3.845\%                     & 3.595\%                     & 3.659\%                     &  136.16 & 162.85 & 145.67 & 143.40\\
L3   & 3.629\%                  & 3.845\%                     & 3.585\%                     & 3.663\%                     &  136.15 & 162.08 & 144.72 & 143.05\\
L4   & 3.612\%                  & 3.848\%                     & 3.556\%                     & 3.648\%                     &  136.13 & 163.92 & 143.72 & 143.20\\
L5   & 3.584\%                  & 3.801\%                     & 3.549\%                     & 3.621\%                     &  134.87 & 161.40 & 143.39 & 141.88\\
L6   & 3.524\%                  & 3.720\%                     & 3.510\%                     & 3.561\%                     &  132.12 & 157.31 & 140.91 & 138.92\\
L7   & 3.560\%                  & 3.676\%                     & 3.570\%                     & 3.585\%                     &  131.68 & 154.21 & 141.12 & 138.07\\
L8   & 3.587\%                  & 3.814\%                     & 3.536\%                     & 3.622\%                     &  134.34 & 161.12 & 142.31 & 141.29\\
L9  & 3.562\%                  & 3.771\%                     & 3.495\%                     & 3.590\%                     &  133.83 & 160.86 & 140.67 & 140.61\\
L10  & 3.507\%                  & 3.721\%                     & 3.452\%                     & 3.538\%                     &  131.19 & 157.14 & 138.31 & 137.80\\
L11  & 3.454\%                  & 3.661\%                     & 3.405\%                     & 3.486\%                     &  128.84 & 153.99 & 135.64 & 135.23\\
L12  & 3.533\%                  & 3.759\%                     & 3.483\%                     & 3.568\%                     &  132.03 & 158.37 & 139.69 & 138.83\\
L13  & 3.554\%                  & 3.772\%                     & 3.494\%                     & 3.586\%                     &  132.73 & 158.84 & 140.07 & 139.42\\
L14  & 3.466\%                  & 3.694\%                     & 3.400\%                     & 3.498\%                     &  129.22 & 155.29 & 135.83 & 135.76\\
L15  & 3.492\%                  & 3.744\%                     & 3.437\%                     & 3.531\%                     &  130.21 & 157.32 & 137.43 & 137.08\\
L16  & 3.508\%                  & 3.752\%                     & 3.450\%                     & 3.545\%                     &  130.76 & 157.71 & 137.92 & 137.58\\
L17  & 3.498\%                  & 3.690\%                     & 3.479\%                     & 3.532\%                     &  128.52 & 153.58 & 136.01 & 135.03\\
None MF   & 3.664\%                  & 4.046\%                     & 3.590\%                     & 3.725\%                     &  139.06 & 174.47 & 143.77 & 147.08\\
All locations   & 3.770\%                  & 4.014\%                     & 3.708\%                     & 3.807\%                     &  141.79 & 170.57 & 148.24 & 148.84\\
Average   & 3.563\%                  & 3.693\%                     & 3.555\%                     & 3.587\%                     &  131.72 & 154.81 & 140.33 & 138.06\\
HT   & 3.561\%                  & 3.761\%                     & 3.551\%                     & 3.599\%                      &  133.90 & 159.45 & 143.18 & 140.87\\  \hline
Best & 3.454\%                  & 3.661\%                     & 3.400\%                     & 3.486\%                     &  128.52 &153.58 &135.64 &135.03 \\ \hline
Proposed & 3.395\%                  & 3.582\%                     & 3.026\%                     & 3.359\%                     &   126.09 & 150.47 & 130.02 & 131.75\\ \hline
\end{tabular}
\end{table*}

\begin{table*}[t]
\centering
\caption{Accuracy of different methods in City B}
\label{table2}
\begin{tabular}{lcccccccc}
\hline
     & \multicolumn{1}{l}{MAPE} & \multicolumn{1}{l}{$\text{MAPE}_{\text{Noon}}$} & \multicolumn{1}{l}{$\text{MAPE}_{\text{Night}}$} & \multicolumn{1}{l}{$\text{MAPE}_{\text{Com}}$} & \multicolumn{1}{l}{MAE}       & \multicolumn{1}{l}{$\text{MAE}_{\text{Noon}}$}   & \multicolumn{1}{l}{$\text{MAE}_{\text{Night}}$}    & \multicolumn{1}{l}{$\text{MAE}_{\text{Com}}$}    \\ \hline
L0   & 3.956\%                  & 3.983\%                     & 3.769\%                     & 3.924\%                     &   69.22 & 78.28 & 68.04 & 70.80\\
L1   & 4.523\%                  & 4.754\%                     & 4.652\%                     & 4.595\%                     &   82.64 & 96.34 & 88.76 & 86.60\\
L2   & 4.051\%                  & 4.187\%                     & 4.109\%                     & 4.090\%                     &   72.77 & 83.54 & 76.74 & 75.72\\
L3   & 4.255\%                  & 4.321\%                     & 4.195\%                     & 4.257\%                     &   76.67 & 85.99 & 78.64 & 78.93\\
L4   & 4.312\%                  & 4.383\%                     & 4.279\%                     & 4.320\%                     &   77.67 & 87.35 & 80.43 & 80.16\\
L5   & 4.077\%                  & 4.258\%                     & 4.010\%                     & 4.100\%                     &   73.11 & 84.71 & 74.60 & 75.73\\
L6   & 4.254\%                  & 4.417\%                     & 4.134\%                     & 4.263\%                     &   76.29 & 88.23 & 77.07 & 78.84\\
L7   & 4.541\%                  & 4.780\%                     & 4.633\%                     & 4.607\%                     &   83.04 & 97.04 & 88.40 & 86.91\\
L8   & 4.213\%                  & 4.367\%                     & 4.190\%                     & 4.239\%                     &   76.00 & 87.46 & 78.54 & 78.80\\
L9  & 4.560\%                  & 4.788\%                     & 4.688\%                     & 4.631\%                     &   83.33 & 96.97 & 89.47 & 87.29\\
L10  & 4.109\%                  & 4.293\%                     & 4.038\%                     & 4.132\%                     &   73.84 & 85.76 & 75.55 & 76.56\\
L11  & 3.948\%                  & 4.131\%                     & 3.931\%                     & 3.981\%                     &   70.44 & 81.93 & 72.66 & 73.18\\
L12  & 4.534\%                  & 4.795\%                     & 4.642\%                     & 4.608\%                     &   82.65 & 96.87 & 88.38 & 86.64\\
L13  & 4.562\%                  & 4.805\%                     & 4.618\%                     & 4.622\%                     &   83.34 & 97.38 & 87.96 & 87.07\\
L14  & 4.202\%                  & 4.312\%                     & 4.114\%                     & 4.207\%                     &   75.36 & 85.95 & 76.68 & 77.74\\
L15  & 3.858\%                  & 3.936\%                     & 3.695\%                     & 3.841\%                     &   68.26 & 76.91 & 67.27 & 69.79\\
L16  & 3.885\%                  & 4.057\%                     & 3.746\%                     & 3.892\%                     &   68.70 & 78.55 & 68.61 & 70.65\\
L17  & 3.992\%                  & 4.180\%                     & 3.807\%                     & 3.992\%                     &   71.15 & 82.22 & 69.94 & 73.12\\
L18  & 4.170\%                  & 4.354\%                     & 4.052\%                     & 4.183\%                     &   75.00 & 86.95 & 75.45 & 77.48\\
L19  & 4.348\%                  & 4.613\%                     & 4.254\%                     & 4.382\%                     &   79.45 & 93.49 & 80.60 & 82.49\\
None MF   & 3.964\%                  & 4.014\%                     & 3.935\%                     & 3.968\%                     &   68.93 & 75.61 & 70.62 & 70.61\\
All Grids   & 4.118\%                  & 4.227\%                     & 3.881\%                     & 4.092\%                     &   73.52 & 83.97 & 71.53 & 75.21\\
Average   & 4.251\%                  & 4.429\%                     & 4.253\%                     & 4.287\%                     &   76.71 & 88.68 & 79.86 & 79.74\\ 
HT   & 3.858\%                  & 3.936\%                     & 3.695\%                     & 3.841\%                     &   68.26 & 76.91 & 67.27 & 69.79\\ \hline
Best & 3.858\%                  & 3.936\%                     & 3.695\%                     & 3.841\%                     &   68.26 & 75.61 & 67.27 & 69.79\\ \hline
Proposed & 3.623\%                  & 3.632\%                     & 3.459\%                     & 3.592\%                     &   62.89 & 69.41 & 62.40 & 64.09 \\ \hline
\end{tabular}
\end{table*}

\subsubsection{Analysis}
The mean and standard variance (Std) of different locations in City A and City B are presented in Table \ref{table3}. 
The deviation between the maximum altitude (59m) and minimum altitude (46m) in City A is 13m. In contrast, the deviation between the maximum altitude (1590m) and minimum altitude (-6m) in City B is nearly 1600m. The mountainous terrain in the northern part of City B contributes to a greater variation and lower mean compared to the southern part of the city. The temperature variance among different locations in City B is also larger than in City A, potentially due to the difference in topography. 
It is worth noting that the accuracy improvements of our method in City B are more pronounced than in City A. This phenomenon suggests that our method may yield higher accuracy improvements in regions with significant MF variations caused by terrain differences.

In City A, as shown in Fig. \ref{temperature humidity a}, there is a significant difference in geo-distributed MF during noon and evening, while the differences are less pronounced at other periods. On the other hand, as shown in Fig. \ref{temperature humidity b}, the weather variations between different collection points are evident in almost every period in City B. As a result, the accuracy improvement of 
$\text{MAPE}_{\text{Noon}}$, $\text{MAE}_{\text{Noon}}$, $\text{MAPE}_{\text{Night}}$, $\text{MAE}_{\text{Night}}$ in city A
are much greater than in other periods, while in city B, the accuracy improvement is consistently close and more prominent across most periods.

\begin{table}[t]
\centering
\caption{Mean and Standard Variance of Temperature}
\label{table3}
\begin{tabular}{ccccc}
\hline
    & \multicolumn{2}{c}{City A} & \multicolumn{2}{c}{City B} \\ \hline
    & Mean         & Std         & Mean         & Std         \\ \hline
L0  & 16.470       & 10.424      & 22.349       & 6.413       \\
L1  & 16.430       & 10.414      & 22.570       & 6.522       \\
L2  & 16.377       & 10.414      & 22.499       & 6.438       \\
L3  & 16.338       & 10.390      & 22.400       & 6.489       \\
L4  & 16.492       & 10.558      & 22.292       & 6.508       \\
L5  & 16.466       & 10.549      & 22.241       & 6.551       \\
L6  & 16.416       & 10.534      & 22.299       & 6.780       \\
L7  & 16.368       & 10.512      & 22.469       & 6.680       \\
L8  & 16.324       & 10.482      & 22.419       & 6.576       \\
L9  & 16.310       & 10.636      & 22.302       & 6.602       \\
L10 & 16.355       & 10.644      & 22.230       & 6.605       \\
L11 & 16.271       & 10.633      & 22.151       & 6.589       \\
L12 & 16.177       & 10.595      & 22.258       & 6.833       \\
L13 & 16.145       & 10.537      & 22.238       & 6.759       \\
L14 & 16.184       & 10.717      & 22.173       & 6.720       \\
L15 & 16.133       & 10.698      & 22.188       & 6.699       \\
L16 & 16.106       & 10.682      & 22.089       & 6.831       \\
L17 & 16.069       & 10.799      & 22.054       & 6.818       \\
L18 &    -          &     -       & 22.065       & 6.876       \\
L19 &    -          &      -       & 21.871       & 7.066       \\ \hline
\end{tabular}
\end{table}

\subsection{Model Interpretation}
\subsubsection{City A}
\begin{figure}[t]
\centering
\includegraphics[scale=0.4]{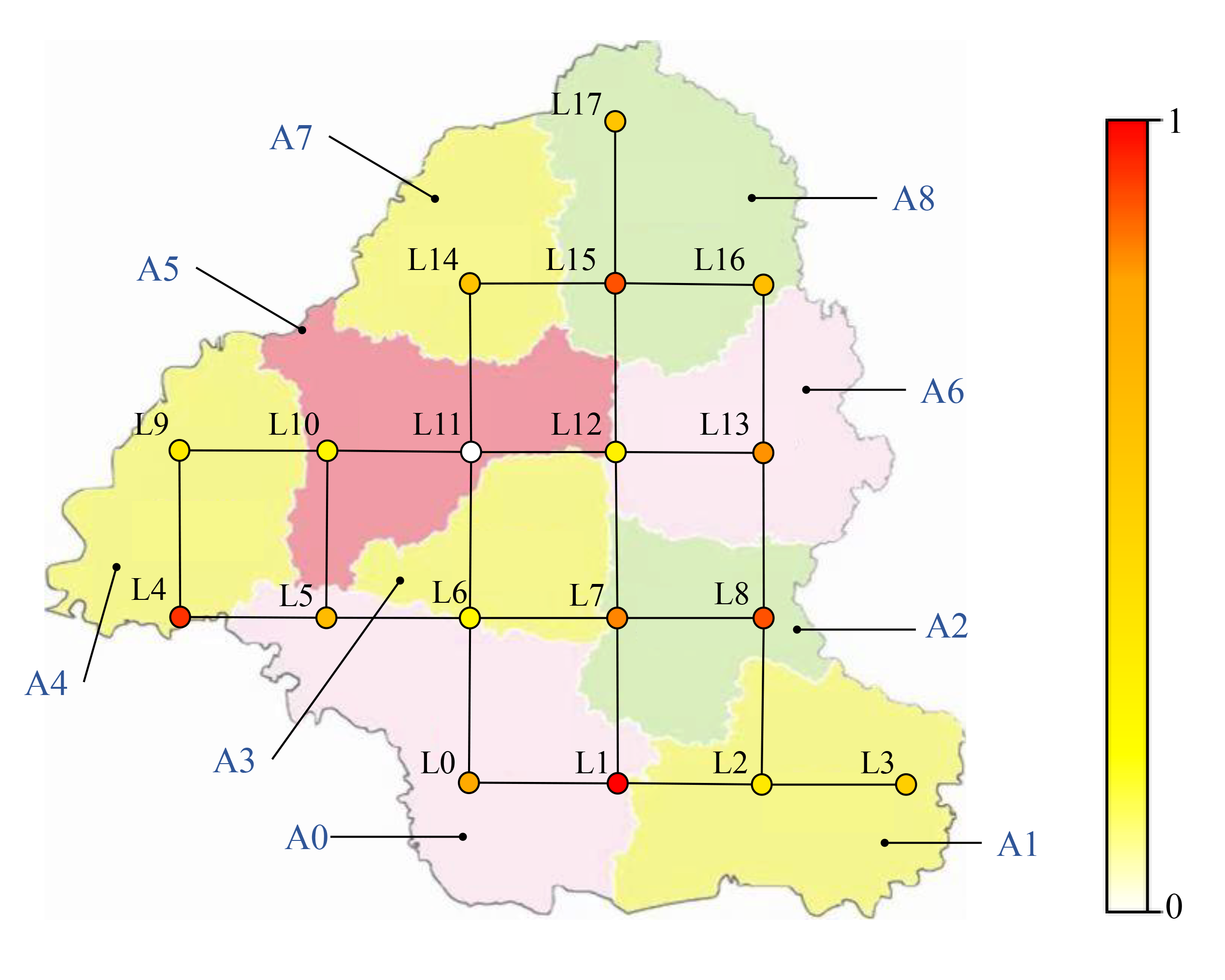}
\caption{Model interpretation result of City A}
\label{shapA}
\end{figure}

The model interpretation result of City A is shown in Fig. \ref{shapA}. The color of the locations represents their importance, with darker shades indicating greater importance of the corresponding grid points. The most important locations for collecting MF data are L1, L4, and L15. City A is divided into nine regions, with A0, A4, and A8 being the top three regions with the highest GDP. Interestingly, we found that L1, L4, and L15 are located within A0, A4, and A8, respectively.
This finding suggests a correlation between the importance of MF from different locations and the economy of the corresponding areas. Areas with higher GDP may tend to have higher electricity loads, which are impacted by MF in the surrounding area. Therefore, the MF in areas with higher GDP becomes more critical for accurate forecasting.

\subsubsection{City B}
\begin{figure}[t]
\centering
\includegraphics[scale=0.42]{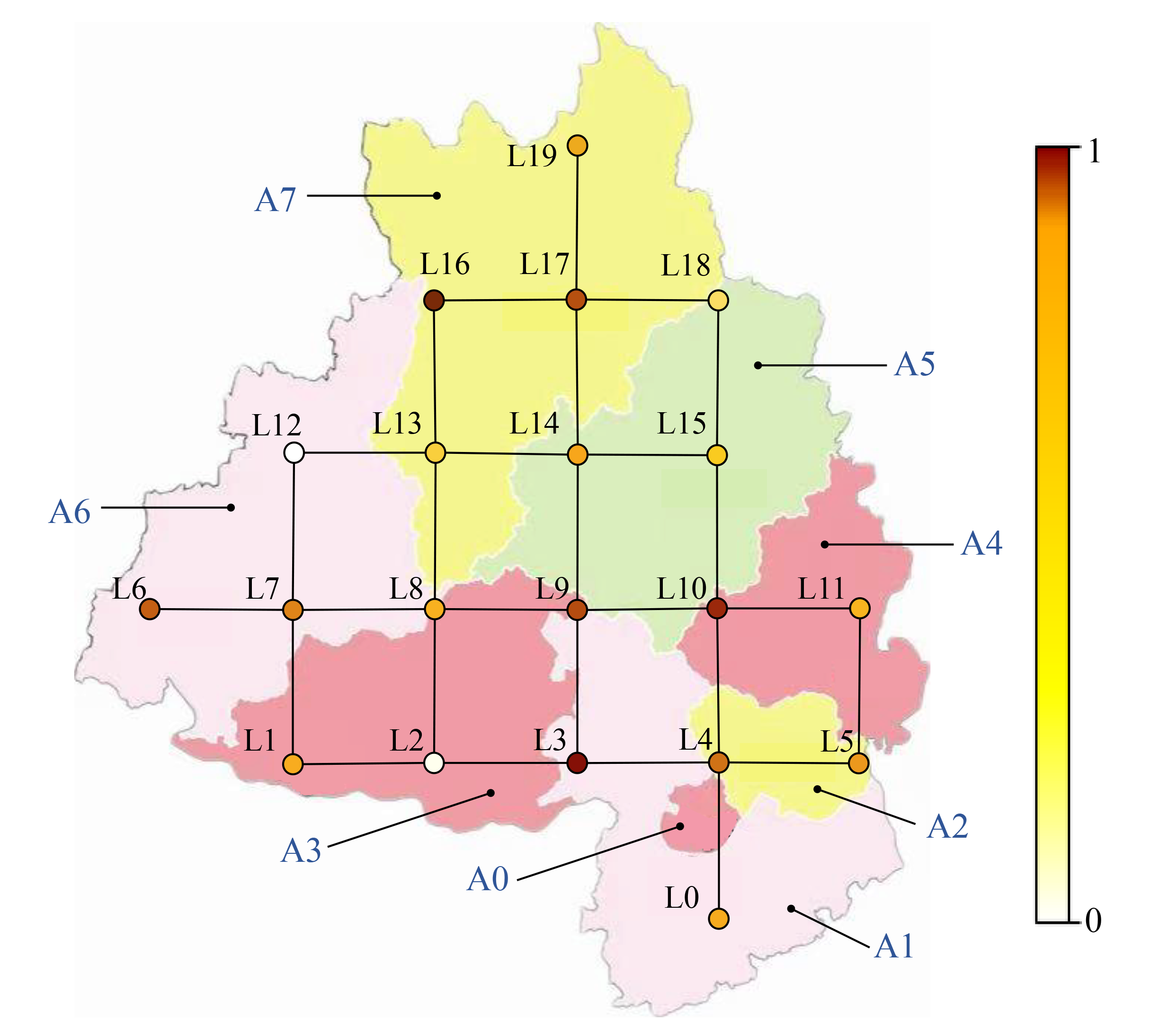}
\caption{Model interpretation result of City B}
\label{shapB}
\end{figure}

The interpretation results of the model for City B are illustrated in Fig. \ref{shapB}. The most important locations for collecting MF data are L3 L10, L16, L17, and L9. City B is divided into eight areas, with A0, A1, A4, and A7 being the top four areas with the highest GDP. Similar to City A, the most crucial MF collection locations are situated within the economically developed regions. Furthermore, although A7 does not have the highest GDP and there is a certain gap in GDP compared to the previous three regions (A0, A1, A4), the mainstay industry of A7 is agriculture. Agriculture is highly sensitive to temperature and humidity, which leads to that the meteorological data collected in A7 has a greater influence on forecasting accuracy. 

\subsubsection{Analysis}
From the interpretation results of the models for City A and City B, 
It can be concluded that there is a strong correlation between the importance of meteorological collection points and the GDP and mainstay industries. 
Additionally, we observed that within the same area, grid points that are further away from the center tend to be more important. For instance, in City A, L1 is more important than L0, and L4 is more important than L9. Similarly, in City B, L3, and L9 are more important than L0. This phenomenon may be attributed to the fact that certain industries, such as manufacturing, often have a preference for locating in suburban areas rather than residential areas. 
These findings can guide us to build more MF collection stations in suburbs of economically developed areas to improve the precision of MF, thereby achieving higher load forecasting accuracy.

\section{Conclusions}
To address the challenge of selecting MF from various locations while considering their spatial relationship, this paper proposes a framework to learn meteorological representation of geo-distributed MF. The graph convolutional neural network is adopted as the representation generator of geo-distributed MF. To further explore the relationship between meteorological factors and electricity load, the representation generator is trained simultaneously with the load forecasting model.
Additionally, this work provides an accelerated Shapley value to the graph-based model based on Monte Carlo sampling and weighted linear regression for model interpretation. We conduct the experiments on the dataset collected from two cities in China. The experiment results show our method can achieve better forecasting accuracy, especially in extreme scenarios such as the ``accumulation temperature effect" in summer and ``sudden temperature change" in winter. The model interpretation results reveal the correlation between the importance of meteorological factors from different locations with the GDP, and mainstay industries. These findings have significant implications for practical guidance, providing valuable insights for deploying weather stations.

\label{Conclusions}

\appendices
\ifCLASSOPTIONcaptionsoff
  \newpage
\fi

\end{document}